\documentclass{article}

\usepackage{arxiv}

\usepackage[utf8]{inputenc} % allow utf-8 input
\usepackage[T1]{fontenc}    % use 8-bit T1 fonts
\usepackage{hyperref}       % hyperlinks
\usepackage{url}            % simple URL typesetting
\usepackage{booktabs}       % professional-quality tables
\usepackage{amsfonts}       % blackboard math symbols
\usepackage{nicefrac}       % compact symbols for 1/2, etc.
\usepackage{microtype}      % microtypography
\usepackage{amsmath}
\usepackage{cleveref}       % smart cross-referencing
\usepackage{lipsum}         % Can be removed after putting your text content
\usepackage{graphicx}
\usepackage{subcaption}
\usepackage{natbib}
\usepackage{doi}

\usepackage{amssymb}% http://ctan.org/pkg/amssymb
\usepackage{pifont}% http://ctan.org/pkg/pifont
\usepackage{xcolor}
\usepackage{multicol}
\usepackage{multirow}
\usepackage{algorithm} 
\usepackage{algpseudocode} 
\usepackage{tcolorbox}
\usepackage[super]{nth}

\title{Spectral Clustering in Convex and Constrained Settings}

\newif\ifuniqueAffiliation
\uniqueAffiliationtrue

\ifuniqueAffiliation % Standard variant of author block
\author{ {Swarup Ranjan Behera}\thanks{This work constituted my M.Tech Thesis, completed in 2015 at the Department of CSE, Indian Institute of Technology Guwahati.}\\
	Department of Computer Science and Engineering\\
	Indian Institute of Technology Guwahati, India\\
	\texttt{swarupranjanbehera@gmail.com} \\
	\And
	{Vijaya V. Saradhi} \\
	Department of Computer Science and Engineering\\
	Indian Institute of Technology Guwahati, India\\
	\texttt{saradhi@iitg.ac.in} \\
}
\else
\usepackage{authblk}

\newbox{\orcid}\sbox{\orcid}{\includegraphics[scale=0.06]{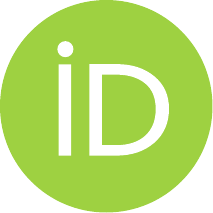}} 
\author[1]{%
	\href{https://orcid.org/0000-0000-0000-0000}{\usebox{\orcid}\hspace{1mm}David S.~Hippocampus\thanks{\texttt{hippo@cs.cranberry-lemon.edu}}}%
}
\author[1,2]{%
	\href{https://orcid.org/0000-0000-0000-0000}{\usebox{\orcid}\hspace{1mm}Elias D.~Striatum\thanks{\texttt{stariate@ee.mount-sheikh.edu}}}%
}
\affil[1]{Department of Computer Science, Cranberry-Lemon University, Pittsburgh, PA 15213}
\affil[2]{Department of Electrical Engineering, Mount-Sheikh University, Santa Narimana, Levand}
\fi

\renewcommand{\shorttitle}{Spectral Clustering in Convex and Constrained Settings}

\hypersetup{
pdftitle={A template for the arxiv style},
pdfsubject={q-bio.NC, q-bio.QM},
pdfauthor={David S.~Hippocampus, Elias D.~Striatum},
pdfkeywords={First keyword, Second keyword, More},
}

\begin{document}
\maketitle

\begin{abstract}
Spectral clustering methods have gained widespread recognition for their effectiveness in clustering high-dimensional data. Among these techniques, constrained spectral clustering has emerged as a prominent approach, demonstrating enhanced performance by integrating pairwise constraints. However, the application of such constraints to semidefinite spectral clustering, a variant that leverages semidefinite programming to optimize clustering objectives, remains largely unexplored. In this paper, we introduce a novel framework for seamlessly integrating pairwise constraints into semidefinite spectral clustering. Our methodology systematically extends the capabilities of semidefinite spectral clustering to capture complex data structures, thereby addressing real-world clustering challenges more effectively. Additionally, we extend this framework to encompass both active and self-taught learning scenarios, further enhancing its versatility and applicability. Empirical studies conducted on well-known datasets demonstrate the superiority of our proposed framework over existing spectral clustering methods, showcasing its robustness and scalability across diverse datasets and learning settings. By bridging the gap between constrained learning and semidefinite spectral clustering, our work contributes to the advancement of spectral clustering techniques, offering researchers and practitioners a versatile tool for addressing complex clustering challenges in various real-world applications. Access to the data, code, and experimental results is provided for further exploration (\url{https://github.com/swarupbehera/SCCCS}).
\end{abstract}

% keywords can be removed

\section{Introduction}  \label{sec:intro}

Clustering, a foundational task in data analysis, entails the segmentation of data points to delineate groups characterized by internal coherence and external dissimilarity. Widely explored and applied across diverse domains of machine learning, clustering bears structural resemblance to the graph partitioning problem, owing to its analogous node-edge configuration mirroring the entity-relation structure ingrained within datasets~\citep{Luxburg:2007:TSC:1288822.1288832}. Herein, each graph node corresponds to a data instance, and each edge denotes the relationship between two such nodes.

The graph partitioning problem is acknowledged as NP-hard. Spectral Clustering (SC)~\citep{conf/nips/NgJW01}, a prominent method employed to address this challenge, employs spectral relaxation, decoupled from direct optimization, thereby complicating the quest for globally optimal clustering outcomes. To mitigate this, semidefinite relaxation is harnessed within SC, yielding convex optimization~\citep{Boyd:2004:CO:993483}. This refined approach, termed Semidefinite Spectral Clustering (SDSC)~\citep{Kim20062025}, aims to ameliorate the inherent optimization intricacies prevalent in conventional SC methodologies.

In practical applications, domain experts often harbor invaluable background knowledge pertaining to datasets, which can substantially boost clustering. Such knowledge is commonly expressed as constraints and seamlessly woven into existing clustering frameworks, giving rise to Constrained Clustering methodologies~\citep{Wagstaff01constrainedk-means}. Constraints are incorporated into SC, with three distinct methodologies emerging based on the selection process. The first, known as Constrained Spectral Clustering (CSC)~\citep{Wang:2010:FCS:1835804.1835877}, randomly selects constraints. The second method, termed Active Spectral Clustering (ASC)~\citep{5694010}, involves the active or incremental selection of constraints. Lastly, Self Taught Spectral Clustering (STSC)~\citep{doi:10.1137/1.9781611973440.48} is characterized by constraints that are self-taught. While considerable research has been devoted to constraint integration within SC, hardly any attention has been directed towards its integration within SDSC.

This paper introduces three formulations aimed at extending SC to incorporate convex and constrained settings. Our primary goal is to seamlessly integrate pairwise constraints into SDSC, demonstrating its effectiveness across real-world datasets. Our contributions include:
    
    \begin{itemize}
        \item Introducing a framework for embedding pairwise constraints into SDSC.
        \item Proposing an extension from passive to active learning paradigms by actively selecting informative constraints, particularly beneficial in sequential scenarios such as video stream clustering.
        \item Further extending the framework to self-taught learning, allowing for the self-derivation of constraints without human intervention, providing a viable solution in constraint-scarce environments.
    \end{itemize}

The paper is structured as follows: Section~\ref{sec:bg} provides background information on different clustering methods. Section~\ref{sec:pf} introduces the proposed frameworks. In Section \ref{sec:er}, we explore the experimental setup and datasets utilized, followed by the presentation of results and analysis. The paper culminates with a discussion and considerations for future directions in Section~\ref{sec:c}.

\section{Background} \label{sec:bg}

\begin{figure}[!tbh]
    	\centering
    	\includegraphics[scale=0.1]{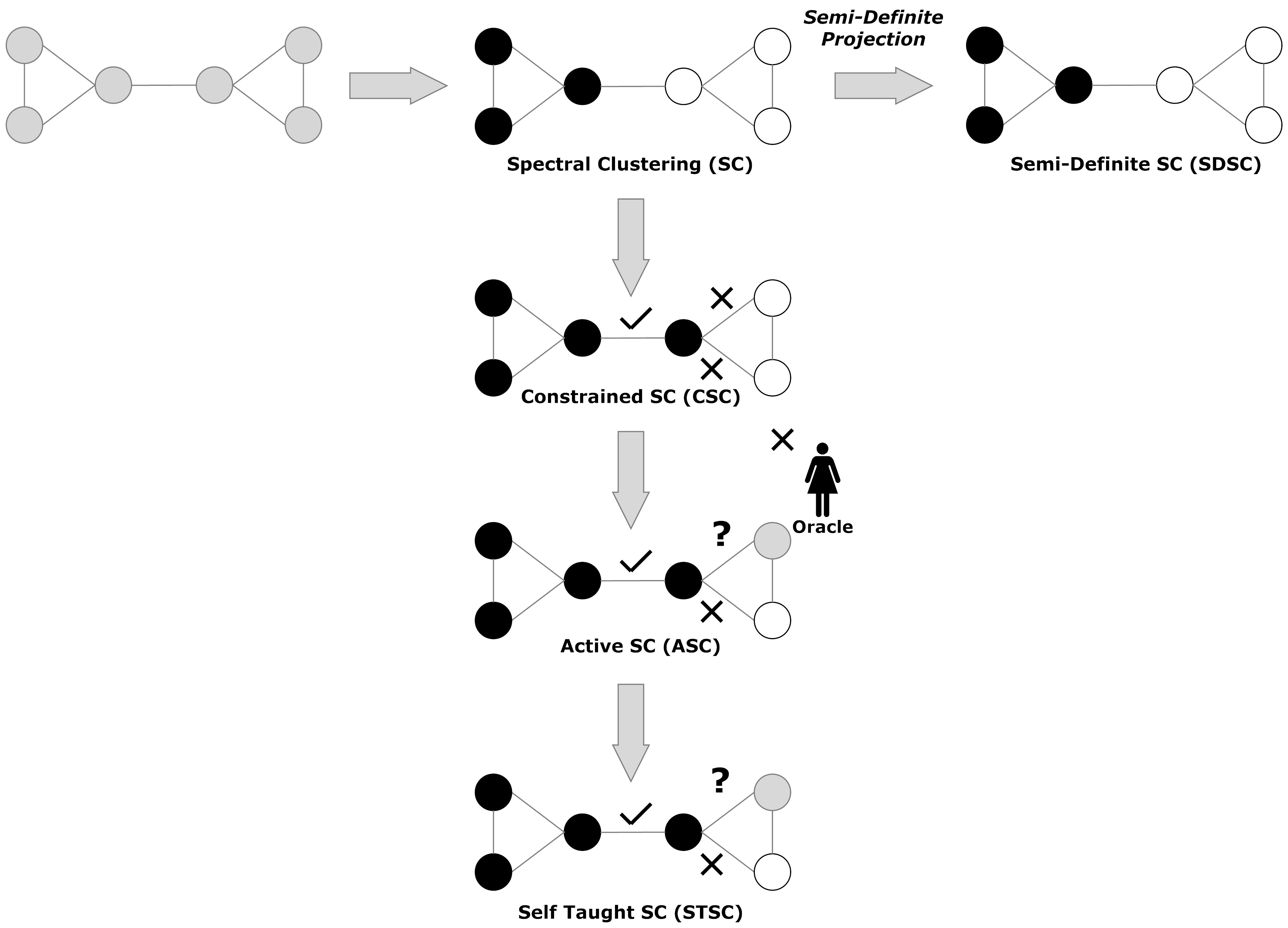}
    	\caption{Overview of data clustering methods.}
    	\label{sa}
    \end{figure}

To provide context for this paper, we commence with a visual depiction of prevalent data clustering methods in Figure~\ref{sa}. Let $G(V,E)$ denote a connected graph, where $V=\{x_i\}$ represents the set of $n$ vertices corresponding to data points and $E=\{w_{ij}\}$ denotes the set of weighted edges indicating pairwise similarity values. The graph $G(V,E)$ can be divided into $k$ disjoint sets (we consider multi-way graph equipartitioning), $A_1, ..., A_k$, with $|A_i|=m$ for $i=1,2,...,k$, satisfying $A_1 \cup A_2 \cup ... \cup A_k = V$ ($|V|=n$, i.e., $n=m \times k$) and $A_1 \cap A_2 \cap ... \cap A_k = \emptyset$ by removing inter set edges. The sum of weights of the removed edges represents the degree of dissimilarity between these $k$ sets, known as the cut value. A lower cut value indicates a superior partitioning.

The cut criterion for the multi-way graph equipartitioning problem is defined as follows:

    \begin{equation}	\label{eq1}
    \begin{split}
    arg\: _{X} \: min \;  \;  \;  \;  	&  tr\left ( X^{T}LX \right ) \\
    s.t. \;  \;  \;  \;  & \text{X is an indicator matrix}  \\ 
    & \text{Each node is assigned to atleast one cluster} \\ 
    & \text{Cluster sizes are identical} 
    \end{split}
    \end{equation}

Here, $X$ represents the partition matrix of dimension $n\times k$, given as $X =\left [ \vec{x_{1}} \cdots \vec{x_{k}} \right ]$, where $\vec{x_{i}}$ are indicator vectors representing partitions. The indicator vector $1_A = {\left ( f_1,...,f_n \right )}' \in \mathbb{R}^n $ is defined such that its entries $f_i = 1$ if $v_i \in A$, and $f_i = 0$ otherwise. The Laplacian matrix $L$ is calculated as $L = D - W$, where $W$(=$[w_{ij}] \in \mathbb{R}^{n\times n}$) represents the adjacency or similarity matrix, and $D$ is a diagonal matrix with the degrees $d_1,...,d_n$ on the diagonal, where $d_i=\sum_{j \in V}w_{ij}$ denotes the degree of node $i$. The optimization is minimized when $X$ forms an orthogonal basis for the subspace spanned by the eigenvectors corresponding to the $k$ smallest eigenvalues of $L$.
    
The multi-way graph equipartitioning problem poses a challenge due to its NP-hard and non-convex nature. Spectral relaxation serves as a prevalent method for tackling such problems, commonly known as Spectral Clustering (SC). While various SC techniques exist, they generally follow a similar foundational framework. For k-way clustering, SC typically begins by identifying the top $k$ eigenvectors of the Graph Laplacian matrix L, followed by running k-means clustering on the normalized rows of these eigenvectors. SC tends to perform optimally when the graph Laplacian matrix exhibits a well-defined block diagonal structure, indicating clear separation among different sub-clusters. However, real-world datasets often contain noise or artifacts, making it challenging to achieve such idealized structures. Consequently, SC struggles to yield globally optimal clustering results due to its indirect relationship with optimization, relying instead on heuristic approaches.

Employing convex optimization enhances the performance of data clustering methods. By leveraging semidefinite relaxation, it becomes feasible to attain a block diagonal structure within the Laplacian matrix, leading to the formulation known as Semidefinite Spectral Clustering (SDSC)~\citep{Kim20062025}. Equation~\ref{eq1} can be reformulated as a constrained quadratic programming problem to facilitate semidefinite relaxation, as follows.

\begin{equation}  \label{eq2}
\begin{split}
arg\: _{X} \: min \;  \;  \;  \;  	&  tr\left ( X^{T}LX \right ) \\
s.t. \;  \;  \;  \;  &  X\circ X= X \;  \;  \;  \; \;  \;  \;  \Leftrightarrow \left \{ X:x_{ij}^{2}=x_{ij}, \forall i,j \right \} \\
& X\vec{e}_{k}= \vec{e}_{n} \\ 
& X^{T}\vec{e}_{n}= m\vec{e}_{k}  \;  \;  \;  \;  \Leftrightarrow \left \{X: \left \| X\vec{e}_{k}-\vec{e}_{n} \right \|^{2}+\left \| X^{T}\vec{e}_{n}- m\vec{e}_{k} \right \| ^{2}= 0 \right \}
\end{split}
\end{equation}

Here, the operator $\circ$ denotes the Hadamard product, also known as the element-wise product, and $\vec{e}_{i}$ represents the indicator vector. SDSC first finds the optimal feasible matrix through the projected semidefinite relaxation and forms the optimal partition matrix from it. It then runs k-means clustering on the normalized rows of the eigen vectors of optimal feasible matrix.

Clustering traditionally addresses unsupervised problems, yet there are scenarios where unsupervised clustering alone proves insufficient. Fortunately, in real-world applications, experimenters often possess domain-specific background knowledge or data insights that could enhance the clustering process. Constrained clustering can be viewed as the integration of domain knowledge or constraints into the fundamental clustering framework. These constraints may manifest as domain knowledge, such as class labels or instance-level information \citep{wc-cic-00}. While obtaining class labels can be laborious and time-consuming for human experts, acquiring instance-level constraints or pairwise relations is often more feasible. Pairwise constraints convey a priori knowledge regarding which instances should or should not be grouped together. These constraints typically consist of two types: Must Link (ML) constraints, indicating instances that must be placed in the same cluster, and Cannot Link (CL) constraints, denoting instances that must not belong to the same cluster. \cite{Wang:2010:FCS:1835804.1835877} introduced a more flexible and principled framework known as Flexible Constrained Spectral Clustering (FCFS) to incorporate constraints into SC. Integration of pairwise constraints is achieved within the objective function, while preserving the original graph Laplacian matrix and explicitly encoding the constraints in an $n \times n$ constraint matrix Q, as follows.

\begin{center}
	$ Q_{ij} = Q_{ji} = \left\{\begin{matrix}
	-1 & if \;\; (x_i,x_j) \in CL\\ 
	+1 & if \;\; (x_i,x_j) \in ML \\ 
	0 & otherwise 
	\end{matrix}\right.$\\
\end{center}

Upon encoding the constraints, the cut criterion is transformed into,

\begin{equation}	\label{eq3}
\begin{split}
arg\: _{X} \: min \;  \;  \;  \;  	&  tr\left ( X^{T}LX \right ) \\
s.t. \;  \;  \;  \;  & \text{X is an indicator matrix}  \\ 
& \text{Each node is assigned to atleast one cluster} \\ 
& \text{Cluster sizes are identical}  \\ 
& \text{Lower bound how well the constraints in constraint matrix are satisfied} 
\end{split}
\end{equation}

Optimizing the performance of constrained clustering involves actively selecting the most informative constraints, assuming the presence of a human expert or \textit{Oracle} capable of providing ground truth answers for unknown pairwise relations~\citep{basu:sdm04}. The objective is to enhance the quality of the resulting clustering while minimizing the number of queries made. This approach proves particularly advantageous for problems with a sequential aspect, such as clustering a stream of video data, where specifying constraints incrementally yields more meaningful results than providing them all at once. The active learning framework for constrained spectral clustering, Active Spectral Clustering (ASC)~\citep{5694010}, encompasses two essential components: a constrained spectral clustering algorithm and a query strategy for determining the next best constraint to query. Initially, ASC computes the graph Laplacian matrix L and the Ncut value of the graph using the constrained spectral clustering algorithm. If the Ncut value converges to the ground truth cluster assignment, the clustering result is outputted; otherwise, the framework selects the best constraint to query next based on the query strategy. This iterative process continues until the Ncut value converges to the ground truth cluster assignment.

The performance of constrained clustering can be further enhanced by extending it to the self-teaching setting~\citep{doi:10.1137/1.9781611973440.48}, particularly beneficial in scenarios where expert guidance is limited and consulting an Oracle is impractical. In self-teaching, the existing constraint set for CSC algorithms is enriched by incorporating self-learning mechanisms. This process can significantly improve performance without additional human intervention. Moreover, even when an Oracle is accessible, self-teaching can reduce the number of queries and alleviate the burden on human experts. A systematic framework has been devised by \cite{doi:10.1137/1.9781611973440.48} to leverage both the affinity structure of the graph and the low-rank property of the constraint matrix for augmenting the given constraint set.

\section{Proposed Frameworks} \label{sec:pf}

\begin{figure}[!h]
	\centering
	\includegraphics[scale=0.1]{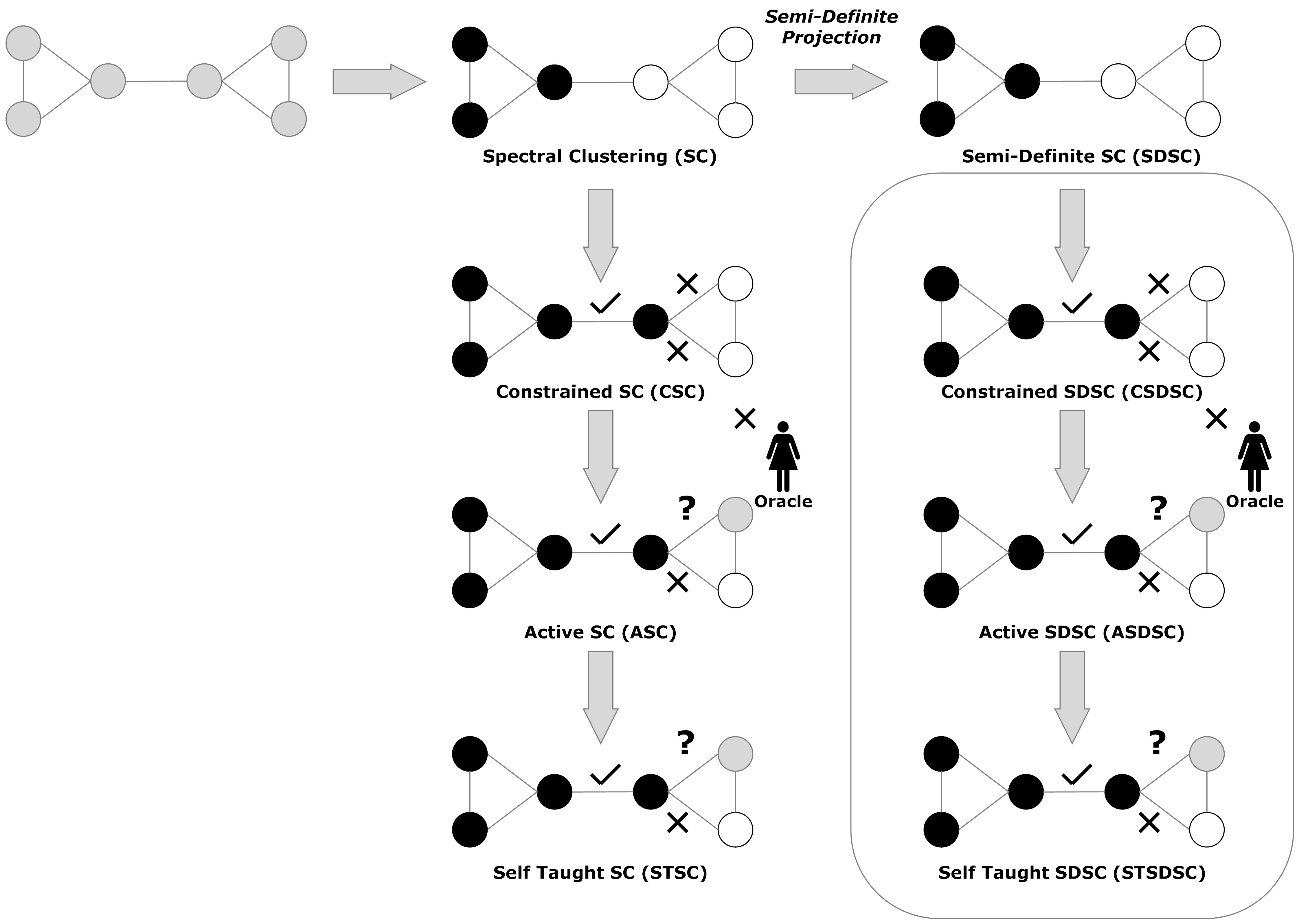}
	\caption{Overview of the proposed frameworks.}
	\label{sa1}
\end{figure}

Constraints have been successfully integrated into various popular clustering algorithms, including k-means clustering, hierarchical clustering, and SC. However, to date, no efforts have been made to incorporate constraints into SDSC. In this work, we leverage a flexible and generalized framework known as FCSC~\citep{Wang:2010:FCS:1835804.1835877} to incorporate constraints into SDSC formulations. On the flip side, we enhance FCSC by exploiting the fact that the eigenvectors of the optimal feasible matrix obtained from SDSC closely resemble piecewise-constant vectors. This suggests that clustering based on these vectors yields successful grouping and outperforms traditional graph Laplacian approaches. Furthermore, we extend our formulation to active and self-taught learning settings. Figure~\ref{sa1} provides a visual overview of our approach. Specifically, we propose three frameworks: Constrained Semidefinite Spectral Clustering (CSDSC), Active Semidefinite Spectral Clustering (ASDSC), and Self-Taught Semidefinite Spectral Clustering (STSDSC).

\subsection {Constrained Semidefinite Spectral Clustering (CSDSC)}

In the formulation of CSDSC, we initially derive the optimal feasible matrix $\tilde{Y}^{*}$ using the projected semidefinite relaxation method proposed by~\cite{Kim20062025}. Subsequently, we normalize $\tilde{Y}^{*}$ and utilize it instead of the Graph Laplacian within the FCSC framework. Here, we incorporate the constraints into the objective function and transform it into a generalized eigenvalue system, which can be deterministically solved in polynomial time. To ensure the identification of ${K-1}$ feasible eigenvectors, we establish a threshold $\beta$ such that $\beta < \lambda _{K-1} vol$, where $\lambda _{K-1}$ denotes the $(K-1)^{th}$ largest eigenvalue of $\bar{Q}$. From the set of feasible eigenvectors, we select the top ${K-1}$ vectors that minimize $ V^T \tilde{Y}^{*}V $, where V represents the set of vertices or data points. Let these ${K-1}$ eigenvectors form the columns of $V\in R^{N\times\left ( K-1 \right ) }$. Finally, we conduct K-means clustering on the rows of V to obtain the final clustering result. The outline of the CSDSC algorithm is presented in Algorithm~\ref{CSDSC}.

\begin{algorithm}[tb]
    \caption{Constrained Semidefinite Spectral Clustering (CSDSC)}\label{CSDSC}
    \begin{algorithmic}[1]
        \Require
        Data points, denoted as vertices $\mathit{V}=\{v_{1},...,v_{n}\}$
        \State Construct the graph Laplacian $L = W - D$, where $W$ is the affinity matrix defined by $W_{ij} = \exp\left\{-\left(\| v_{i}-v_{j} \|^{2}\right)/2\sigma ^{2}\right\}$, and $D$ is the diagonal matrix represented as $D_{ii} = \sum_{j}W_{ij}$
        \State Find the optimal feasible matrix $Y^{*}\in \varphi _{+}^{nk+1}$ by solving the projected semidefinite relaxation using the interior point method, where $k$ is the number of clusters
        \State Form $\tilde{Y}^{*}$, defined as the sum of diagonal blocks of $Y^{*}_{2:nk+1,2:nk+1}$
        \State Construct the constraint matrix $Q$ and normalize it as $\bar{Q} \leftarrow D^{-\frac{1}{2}}QD^{-\frac{1}{2}}$
        \State Compute $\lambda _{k-1}$, the largest eigenvalue of $\bar{Q}$
        \If{$\beta \geq \lambda _{k-1} vol$}
            \State \Return $u^{*} = 0$, where $\beta$ is the threshold
        \Else
            \State Solve the generalized eigenvalue system: $\tilde{Y}^{*}V = \lambda \left( \bar{Q}- \frac{\beta }{vol}\mathbb{I}\right)V$
            \State Remove eigenvectors associated with non-positive eigenvalues and normalize the rest: $V \leftarrow \frac{V}{\| V \|}\sqrt{vol}$
            \State $ V^{*} \leftarrow  \underset{V \in R^{N\times(K-1)}}{\text{arg min}} \:  \text{trace}\left( V^T \tilde{Y}^{*}V  \right) $, where the columns of $V$ are a subset of the feasible eigenvectors generated in the previous step
            \State Renormalize each row of $V^{*}$ to obtain $\left [\tilde{V}*\right ]_{ij}$, i.e., divide $V_{ij}^{*}$ by $\left\{ \sum_{j}\left( V_{ij}*\right)^{2}\right\}^{1/2}$
            \State \Return $U^{*} \leftarrow \text{kmeans}\left( D^{-1/2}\tilde{V}*, \; K \right)$
        \EndIf
    \end{algorithmic} 
\end{algorithm}

\subsection{Active Semidefinite Spectral Clustering (ASDSC)}

Active Semidefinite Spectral Clustering (ASDSC) represents the active learning extension of CSDSC. Similar to the CSDSC algorithm, ASDSC utilizes $\tilde{Y}^{*}$ instead of the Graph Laplacian within the ASC framework \citep{5694010}. Subsequently, we compute the Normalized Cut (Ncut) of the graph $G$ using the Constrained Spectral Clustering algorithm A, which is characterized by the following objective function:

\begin{equation} 	\label{eq4}
    \begin{aligned}
    & \underset{u\in R^N}{argmin} \; \; u^{T}\tilde{Y}^{*}u \\ 
    & \; \; \; \; \; \;  s.t \; \; \;u^{T}Du=vol\left ( G \right )\\
    &  \; \; \; \; \; \; \; \; \; \; \; \; \; u^{T}Qu \geq \alpha
    \end{aligned}
\end{equation}

Here, $D$ represents the Degree matrix, $\text{vol}(G) = \sum_{i=1}^{N}D_{ii}$, and $u$ denotes the relaxed cluster indicator vector.

The solution is obtained through the eigenvalue problem, $\tilde{Y}^{*}u = \lambda \left ( Q - \alpha \mathbb{I} \right )u$, where $\mathbb{I}$ represents an $N \times N$ identity matrix and $\alpha < \lambda _{\text{max}}$, where $\lambda _{\text{max}}$ denotes the largest eigenvalue of $Q$. If the output of algorithm A converges to the ground truth cluster assignment, we output the clustering result; otherwise, we determine the best constraint to query next using the Query strategy S. This process is repeated until convergence. Further details regarding Query strategy S are provided in Algorithm~\ref{S}. The outline of the ASDSC algorithm is presented in Algorithm~\ref{ASDSC}.

\begin{algorithm}[!h]
    \caption{Query Strategy (S)~\citep{5694010}}\label{S}
    \begin{algorithmic}[1]
        \State Let $P^{(t)}_{ij}$ represent the estimation of the pairwise relation between nodes $i$ and $j$ at time $t$:
        \begin{equation*}
        P^{(t)}_{ij} = u^{(t)}_{i}u^{(t)}_{j}
        \end{equation*}
        
        \State Define $d: \mathbb{R} \times \mathbb{R} \rightarrow \mathbb{R}$ as a distance function that measures the error between the current estimation and the ground truth value:
        \begin{equation*}
        d\left( P^{(t)}_{ij}, Q^{*}_{ij} \right) = \left( P^{(t)}_{ij} - Q^{*}_{ij} \right)^{2}
        \end{equation*}
        
        \State As $Q^{*}_{ij}$ remains unknown until after it is queried, the error cannot be computed directly. Instead, the mathematical expectation of the error is computed over the two possible answers from the oracle:
        \begin{equation*}
        E\left( d\left( P^{(t)}_{ij}, Q^{*}_{ij} \right) \right) = d\left( P^{(t)}_{ij}, 1 \right) \cdot Pr\left( Q^{*}_{ij} = 1 \right) + d\left( P^{(t)}_{ij}, -1 \right) \cdot Pr\left( Q^{*}_{ij} = -1 \right)
        \end{equation*}
        
        \State To estimate $Pr\left( Q^{*}_{ij} = 1 \right)$ and $Pr\left( Q^{*}_{ij} = -1 \right)$ based on available information, treat the current constraint matrix $Q^{(t)}$ as an approximation to $Q^{*}$ with missing values. Use the rank-one approximation of $Q^{(t)}$ to recover the unknown entries in $Q^{*}$. Let $\bar{u}^{(t)}$ be the largest singular vector of $Q^{(t)}$, then $\bar{Q}^{(t)} = \bar{u}^{(t)}\left( \bar{u}^{(t)} \right)^{T}$ is the optimal rank-one approximation to $Q^{(t)}$ in terms of Frobenius norm:
        \begin{equation*}
        \begin{aligned}
        & Pr\left( Q^{*}_{ij} = 1 \right) = Pr\left( Q^{*}_{ij} = 1 | Q^{(t)} \right) = \frac{1 + \min\left\{ 1, \max\left\{ -1, \bar{Q}^{(t)}_{ij} \right\} \right\}}{2} \\
        & Pr\left( Q^{*}_{ij} = -1 \right) = 1 - Pr\left( Q^{*}_{ij} = 1 \right)
        \end{aligned}
        \end{equation*}
        
        \State Finally, query the entry that has the maximum expected error:
        \begin{equation*}
        S\left( u^{(t)}, Q^{(t)} \right) = \underset{\left( i,j \right) | Q^{(t)}_{ij} = 0}{\text{arg max}} \; \; E\left( d\left( P^{(t)}_{ij}, Q^{*}_{ij} \right) \right)
        \end{equation*}
    \end{algorithmic} 
\end{algorithm}

\begin{algorithm}[tb]
    \caption{Active Semidefinite Spectral Clustering (ASDSC)}\label{ASDSC}
    \begin{algorithmic}[1]
        \Require
        Data points, denoted as vertices $\mathit{V}=\{v_{1},...,v_{n}\}$
        \State Form the graph Laplacian matrix $L = W - D$, where $W$ represents the affinity matrix defined by $W_{ij} = \exp\left\{-\left(\| v_{i}-v_{j} \|^{2}\right)/2\sigma ^{2}\right\}$, and $D$ is a diagonal matrix with elements $D_{ii} = \sum_{j}W_{ij}$
        \State Obtain the optimal feasible matrix $Y^{*}\in \varphi _{+}^{nk+1}$ by solving the projected semidefinite relaxation using the interior point method, where $k$ denotes the number of clusters
        \State Define $\tilde{Y}^{*}$ as the sum of diagonal blocks of $Y^{*}_{2:nk+1,2:nk+1}$
        \State Initialize an empty constraint matrix $Q^{\left ( 0 \right )}$ with all entries set to zero
        \State Compute the initial clustering using a constrained spectral clustering algorithm $A$ as follows:
        \begin{equation*}
            u^{\left ( 0 \right )}=A\left (\tilde{Y}^{*},Q^{\left ( 0 \right )} \right ) \text{at iteration 0}
        \end{equation*}
        \begin{equation*}
            u^{\left (t \right )}=A\left (\tilde{Y}^{*},Q^{\left ( t\right )} \right ) \text{at iteration t}
        \end{equation*}
        \State Use a query strategy $S$ to evaluate the current clustering $u^{\left (t \right )}$ and the current constraints $Q^{\left (t \right )}$ to determine the next entry in $Q^*$ to query from the oracle:
        \begin{equation*}
            \left ( i,j \right )\leftarrow S\left ( u^{\left ( t \right )},Q^{\left ( t \right )} \right )
        \end{equation*}
        \State Update $Q^{\left ( t \right )}$ to $Q^{\left ( t+1 \right )}$ by filling in $Q^{\left ( t\right )}_{ij}$ and $Q^{\left ( t\right )}_{ji}$ with the value of $Q^{*}_{ij}$, since the constraint matrix is symmetric
        \State Update the clustering as follows:
        \begin{equation*}
            u^{\left ( t+1 \right )}\leftarrow A\left ( \tilde{Y}^{*},Q^{\left ( t+1 \right )} \right )
        \end{equation*} 
        \State Repeat this iteration until a certain stopping criterion is met.
    \end{algorithmic} 
\end{algorithm}

\subsection{Self Taught Semidefinite Spectral Clustering (STSDSC)}
Self Taught Semidefinite Spectral Clustering (STSDSC) represents a further expansion of CSDSC to integrate self-teaching capabilities. Analogous to the CSDSC algorithm, STSDSC utilizes $\tilde{Y}^{*}$ instead of the Graph Laplacian. STSDSC employs the Fixed Point Continuation Module~\citep{doi:10.1137/1.9781611973440.48}, as outlined in Algorithm~\ref{FPCM}. Interested readers are encouraged to refer to the aforementioned paper for further details. The outline of the STSDSC algorithm is presented in Algorithm~\ref{STSDSC}.

\begin{algorithm}[tb]
    \caption{Fixed Point Continuation Module~\citep{doi:10.1137/1.9781611973440.48}}\label{FPCM}
    \begin{algorithmic}[1]
        \Require Partially observed ground truth $P_{\Omega }\left ( Q^{*} \right )$, Graph cut V, K, Parameters $\mu, \; \alpha, \; \beta$
	\State  Initialize: $\eta \leftarrow $ 0.5,  $\tau  \leftarrow $1.9, ${\mu }'  \leftarrow \eta^{-10} \mu$
	\State Initialize: $Q\leftarrow 0$
	\State $G \leftarrow VV^{T}$
    \Repeat
	% \State \textbf{repeat}
	\State ${\mu }'  \leftarrow max\left ( \mu, {\mu}'\eta  \right ) $
	\Repeat
	\State $Y \leftarrow Q - \tau \left ( P_{\Omega }\left ( Q - Q^{*} \right )- \beta G \right )$
	\State Compute the SVD of Y : $Y \leftarrow U_Y \Sigma_Y V_Y^{T}$
    \For {i = 1, . . . , N}
	\State $\Sigma _{Y_{ii}}\leftarrow max\left ( \Sigma _{Y_{ii}} - \tau {\mu}',0 \right )$
	\EndFor
	\State $Q \leftarrow U_Y \Sigma_Y V_Y^{T}$
	\Until convergence
	\Until ${\mu}' = \mu$
    \State \Return Self-taught constraint matrix Q 
    \end{algorithmic} 
\end{algorithm}

\begin{algorithm}[tb]
    \caption{Self-Taught Semidefinite Spectral Clustering (STSDSC)}\label{STSDSC}
    \begin{algorithmic}[1]
        \Require
        Data points, denoted as vertices $\mathit{V}=\{v_{1},...,v_{n}\}$, Constraint Matrix $Q$,  Parameters  $\mu, \; \alpha,\; \beta$, Partially observed ground truth $P_{\Omega }\left ( Q^{*} \right )$
        \State Construct the graph Laplacian $L = W - D$, where $W$ is the affinity matrix defined by $W_{ij} = \exp\left\{-\left(\| v_{i}-v_{j} \|^{2}\right)/2\sigma ^{2}\right\}$, and $D$ is the diagonal matrix represented as $D_{ii} = \sum_{j}W_{ij}$
        \State Find the optimal feasible matrix $Y^{*}\in \varphi _{+}^{nk+1}$ by solving the projected semidefinite relaxation using the interior point method, where $k$ is the number of clusters
        \State Form $\tilde{Y}^{*}$, defined as the sum of diagonal blocks of $Y^{*}_{2:nk+1,2:nk+1}$
        \State  Initialize V to be the K smallest eigenvectors of $\tilde{Y}^{*}$
    	\Repeat
    	\State  Update Q using Fixed Point Continuation Module (Algorithm~\ref{FPCM})
    	\State  $\bar{\tilde{Y}^{*}} \leftarrow \alpha \tilde{Y}^{*} - \beta Q$
    	\State  Update V to be the K smallest eigenvectors of $\bar{\tilde{Y}^{*}}$
	    \Until convergence
        \State \Return Self-taught constraint matrix Q; Constrained cut V
    \end{algorithmic} 
\end{algorithm}

\section{Experiments and Results} \label{sec:er}
The objective here is to demonstrate that our algorithms outperform their counterparts, even with a smaller number of known constraints. We evaluate the performance of our algorithms using three well-known UCI datasets (Hepatitis, Wine, and Iris) \citep{Lichman:2013} and a toy dataset (Twomoon) created for experimentation purposes. The Twomoon dataset comprises 100 instances with two attributes, divided into two classes. Hepatitis dataset consists of 80 instances with 19 attributes, categorized into two classes. The Iris dataset contains 150 instances with four attributes and three classes. Finally, the Wine dataset comprises 178 instances with 13 attributes, divided into three classes. These datasets offer a range of complexities suitable for evaluating our algorithms' performance.

We aim to demonstrate the effectiveness of our algorithms, namely Constrained Semidefinite Spectral Clustering (CSDSC), Active Semidefinite Spectral Clustering (ASDSC), and Self-Taught Semidefinite Spectral Clustering (STSDSC), by comparing them with existing methods including Spectral Clustering (SC), Semidefinite Spectral Clustering (SDSC), Flexible Constraint Spectral Clustering (FCSC), Active Spectral Clustering (ASC), and Self-Taught Spectral Clustering (STSC).

All algorithms are implemented using MATLAB, with the YALMIP toolbox utilized for modeling and solving convex optimization problems. YALMIP focuses on providing a high-level language for algorithm development while utilizing external solvers for computation. We specifically select the SeDuMi solver for semidefinite programming due to its suitability for datasets containing up to 1500 data points, as well as its non-commercial nature. However, we acknowledge that for datasets exceeding 1500 data points, alternatives such as DSDP or SDPlogDet may be more appropriate.

We utilize the Rand Index (RI)~\citep{RI} criterion to assess the effectiveness of the algorithms. The RI represents the proportion of correct decisions relative to the total number of decisions made. A correct decision indicates that the clustering produced by the algorithm aligns with the target clustering. RI values range from 0 to 1, with 1 indicating a perfect clustering result. For Constrained Clustering algorithms, we augment the constraint set by randomly selecting constraints, ranging from 1 percent to 100 percent. The algorithm performances are evaluated based on the Average Rand Index (ARI), which is averaged over 10 repetitions of the constraint generation process.

\begin{figure*}[!tbh]
	\centering
	\begin{subfigure}[h]{0.485\textwidth}
		\centering
		\includegraphics[width=1\textwidth]{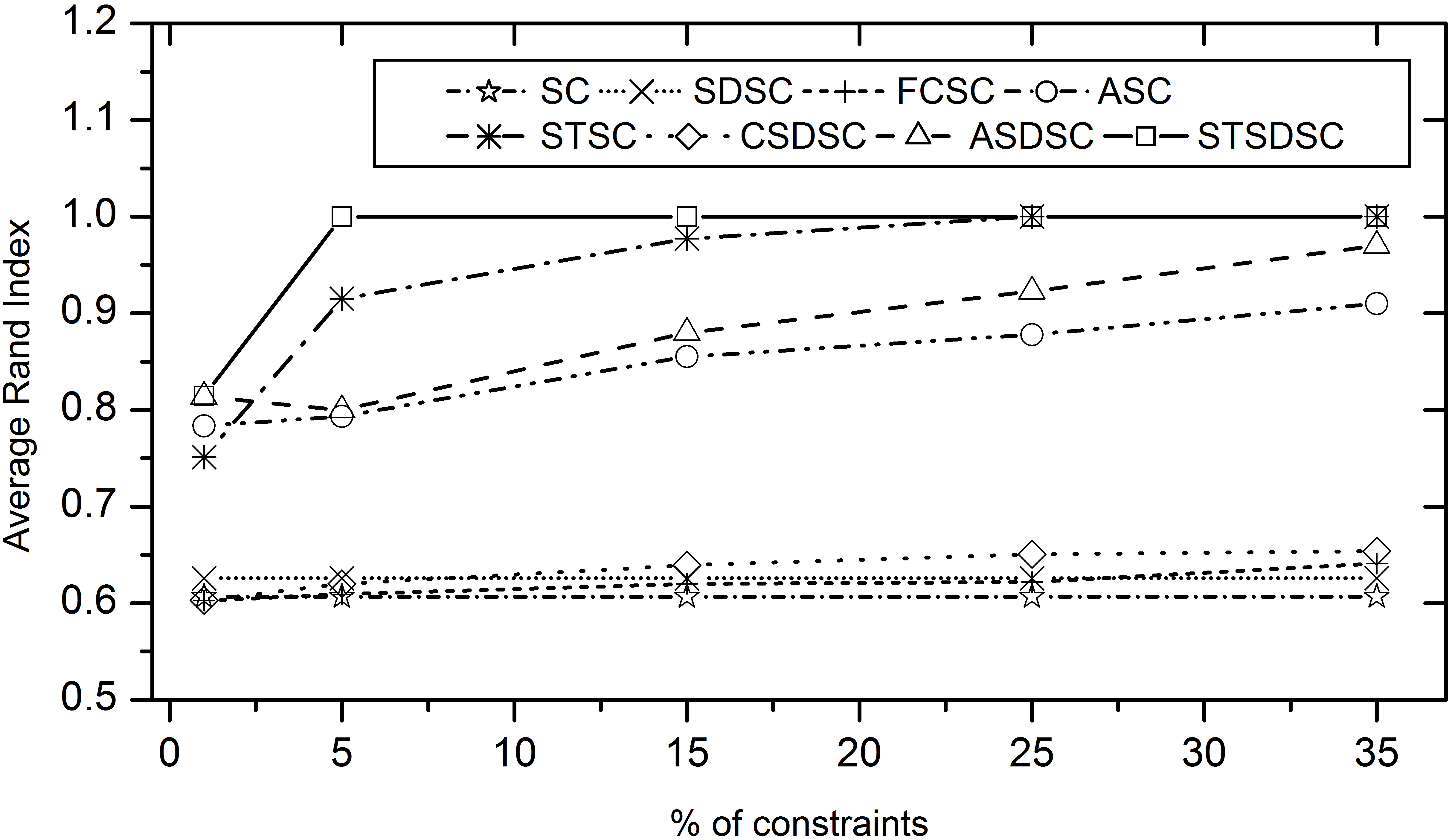}
		\caption{Twomoon dataset}
	\end{subfigure}%
    ~\hspace{1pt}
	\begin{subfigure}[h]{0.485\textwidth}
		\centering
		\includegraphics[width=1\textwidth]{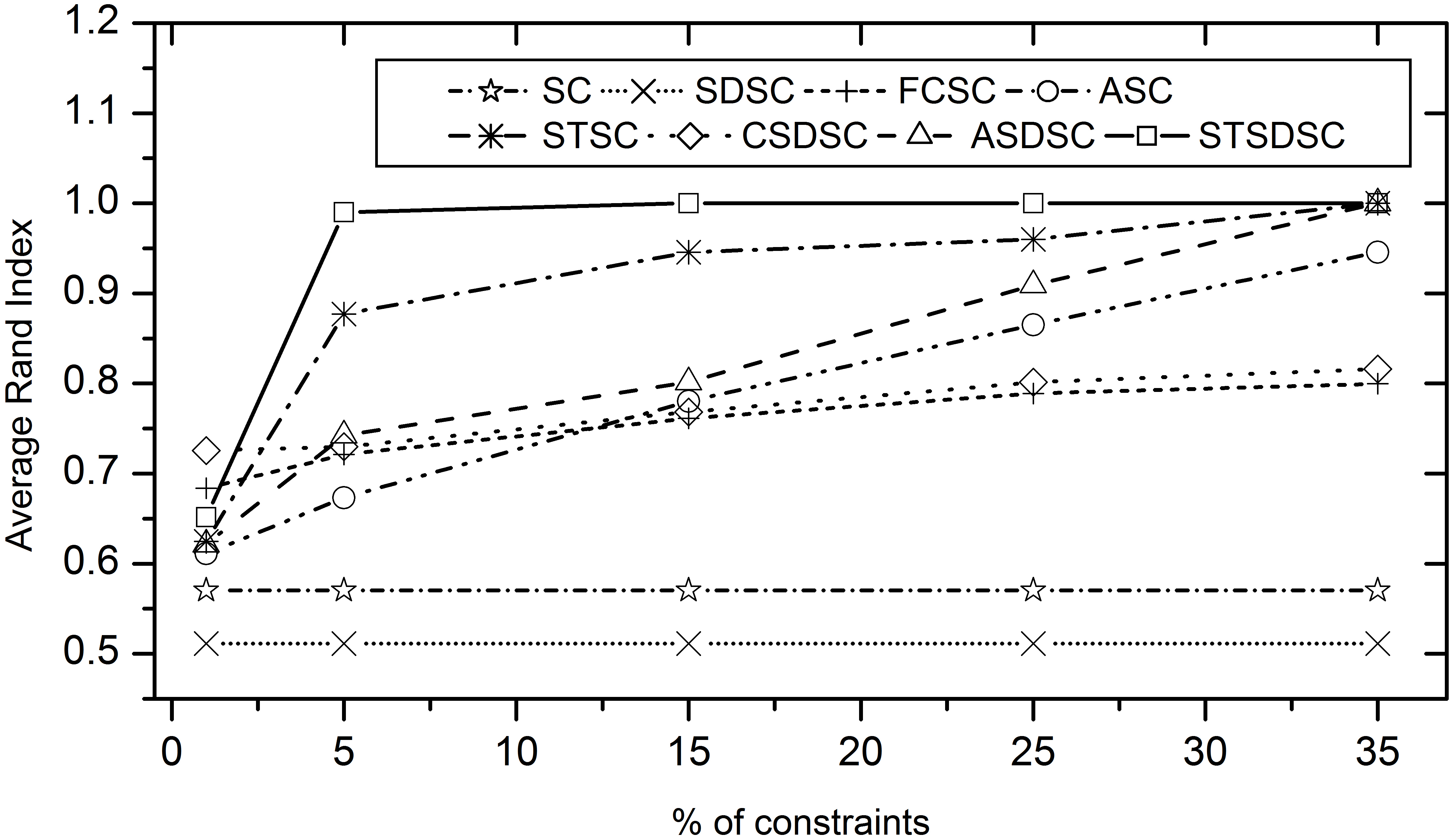}
		\caption{Hepatitis dataset}
	\end{subfigure}

	\begin{subfigure}[h]{0.485\textwidth}
		\centering
		\includegraphics[width=1\textwidth]{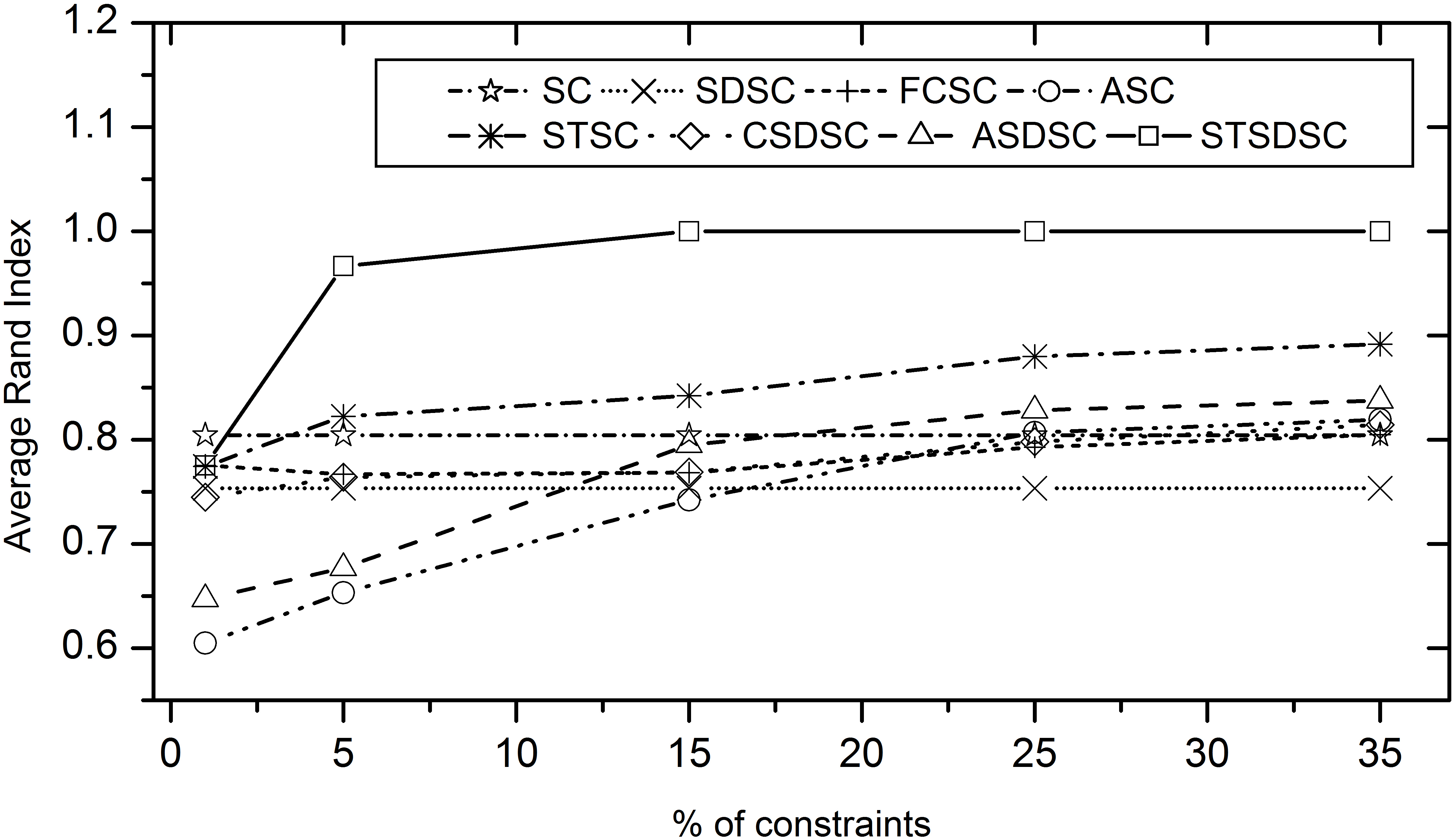}
		\caption{Iris dataset}
	\end{subfigure}%
    ~\hspace{1pt}
	\begin{subfigure}[h]{0.485\textwidth}
		\centering
		\includegraphics[width=1\textwidth]{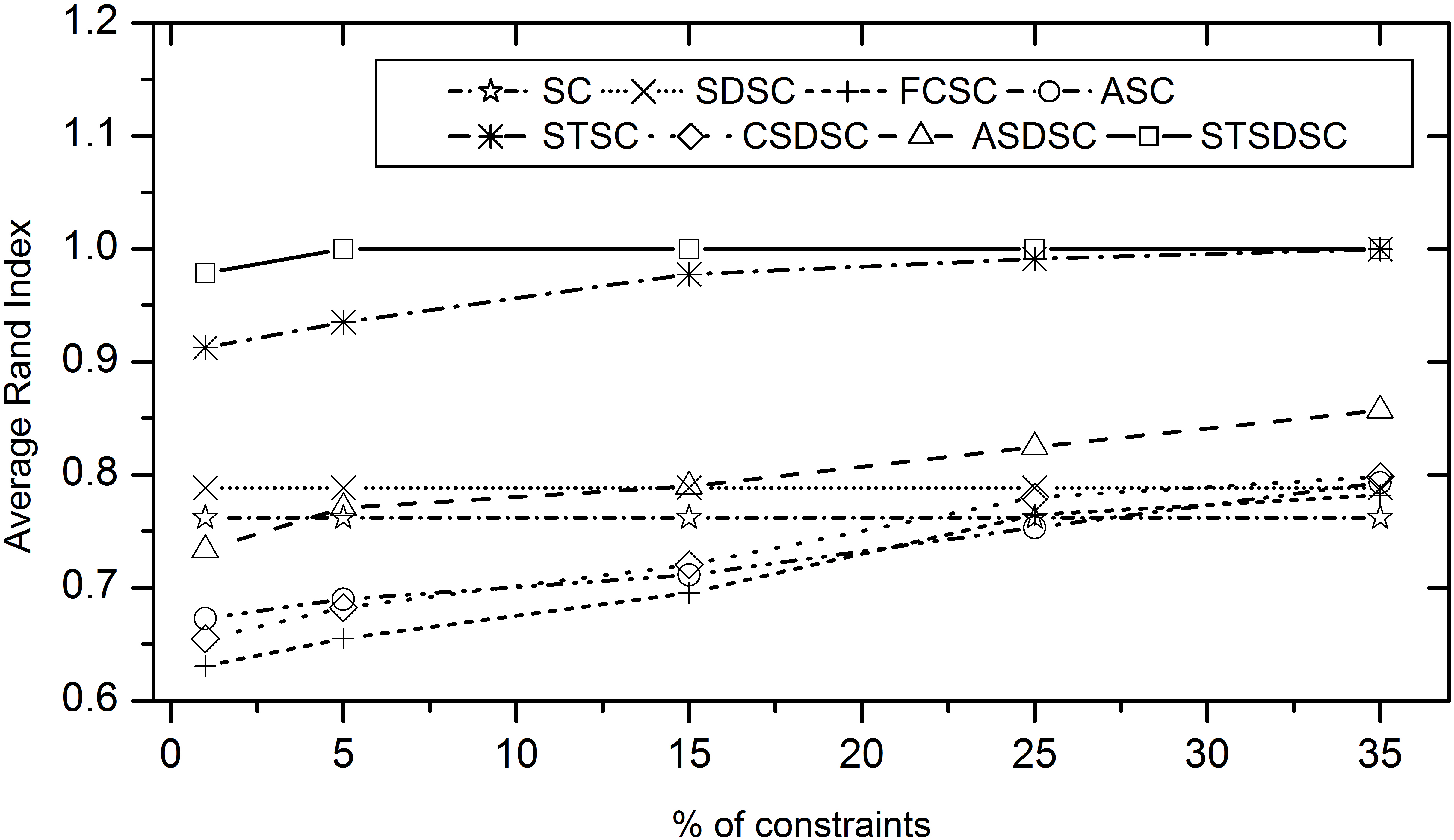}
		\caption{Wine dataset}
	\end{subfigure}
	\caption{Comparison of Average Rand Index among algorithms with varying constraint rates.}
        \label{sa2}
\end{figure*}

The results depicted in Figure~\ref{sa2} provide insights into the performance of various algorithms based on the ARI across different constraint rates. Notably, our CSDSC, ASDSC, and STSDSC algorithms exhibit a tendency to converge to ground truth clustering more rapidly, requiring fewer known constraints compared to other methods. In contrast, SC maintains a consistent ARI regardless of constraint rates, reflecting its independence from such constraints.

An interesting observation arises from the comparison between SDSC and SC. While SDSC outperforms SC in terms of ARI for datasets like Twomoon and Wine, it encounters challenges with the Hepatitis and Iris datasets. This limitation in SDSC's performance has a consequential impact on the effectiveness of our approaches, given that SDSC serves as the foundation for our algorithms.

Further examination reveals that FCSC demonstrates limited performance improvement when compared to active and self-taught constraint selection methods, even when utilizing similarly sized randomly selected constraint sets. ASC exhibits a trend of initial performance degradation, although it tends to recover with the addition of more queried constraints.

In contrast, STSC consistently outperforms other existing approaches across all scenarios. Interestingly, our CSDSC algorithms demonstrate superior performance with increasing constraint rates compared to algorithms without constraints or those with randomly selected constraint settings, albeit they may initially encounter challenges.

ASDSC follows a similar trend to ASC but outperforms other methods except those employing self-taught settings. Notably, STSDSC consistently outperforms other approaches, particularly excelling on the Iris dataset, where alternative methods face significant challenges in the early stages.

\section{Conclusion} \label{sec:c}
In this study, we proposed and evaluated several variants of constrained semidefinite spectral clustering algorithms, namely Constrained Semidefinite Spectral Clustering (CSDSC), Active Semidefinite Spectral Clustering (ASDSC), and Self-Taught Semidefinite Spectral Clustering (STSDSC). Through extensive experiments conducted on various datasets, including Hepatitis, Wine, Iris, and Twomoon, we demonstrated the effectiveness of our proposed algorithms compared to existing methods.

Our results indicate that CSDSC, ASDSC, and STSDSC algorithms outperform traditional spectral clustering and semidefinite spectral clustering approaches, particularly when provided with limited constraints. Moreover, STSDSC consistently outperforms other methods across all datasets, showcasing its robustness and scalability.

Furthermore, our findings highlight the importance of incorporating constraints in spectral clustering algorithms. While random constraint selection strategies show limited improvement, active and self-taught learning settings significantly enhance clustering performance, especially in scenarios with sparse or noisy data.

Overall, our study underscores the significance of constrained semidefinite spectral clustering in addressing real-world clustering challenges. The developed algorithms offer promising avenues for applications in various domains, including image segmentation, social network analysis, and bioinformatics. Future research could focus on extending these methods to handle larger datasets and exploring additional constraint selection strategies to further enhance clustering accuracy.

    \clearpage

\bibliographystyle{unsrtnat}
\bibliography{references}  

\end{document}